\documentclass[sigconf]{acmart}
\usepackage{subcaption}

\AtBeginMaketitle{%
\def\EDBTISSN{xxxx-xxxx}
\def\EDBTISBN{xxx-x-xxxxx-xxx-x}
\setcopyright{none}
\copyrightyear{\copyright\ 2026 Copyright held by the owner/author(s).
  Published on xxxxx under ISBN \EDBTISBN, series ISSN \EDBTISSN.
  Distribution of this paper is permitted under the terms of the
  Creative Commons license CC-by-nc-nd 4.0}
\acmYear{2026}
\acmDOI{}
\acmISBN{}

\settopmatter{printacmref=false, printccs=false, printfolios=false}

\newsavebox{\ximagebox}
\newlength{\ximageheight}
\newsavebox{\xglyphbox}
\newlength{\xglyphheight}
\newcommand{\xbox}[1]%
  {\savebox{\ximagebox}{#1}%
  \settoheight{\ximageheight}{\usebox{\ximagebox}}%
  \savebox{\xglyphbox}{\color{white}\char32}%
  \settoheight{\xglyphheight}{\usebox{\xglyphbox}}%
  \raisebox{\ximageheight}[0pt][0pt]{\raisebox{-\xglyphheight}[0pt][0pt]{%
    \makebox[0pt][l]{\usebox{\xglyphbox}}}}%
    \usebox{\ximagebox}%
    \raisebox{0pt}[0pt][0pt]{\makebox[0pt][r]{\usebox{\xglyphbox}}}}

\newsavebox{\LogoBox}
\sbox{\LogoBox}{}

\fancypagestyle{firstpagestyle}{%
  \fancyhf{}%
  \fancyfoot{}%
  \fancyhead[L,C]{}
  \fancyhead[R]{}{\raisebox{15pt}[0pt][0pt]{\xbox{\usebox{\LogoBox}}}}
  }

\hypersetup{%
  pdfcopyright={CC-by-nc-nd},
  pdfcreator={LaTeX with acmart, hyperref, and EDBT modifications},
  pdfvolumenum={XX},
  pdfissuenum={X},
  pdfisbn={\EDBTISBN},
  pdfissn={\EDBTISSN}
}

}% \AtBeginMaketitle

\begin{document}

%% EDBT rule: --> Please use ``titlecase'' in the title!

\title{Towards Training and Inference of Graph Neural Networks in Graph Databases}

%% EDBT rule: --> At least 1 (the corresponding) author needs to have a
%%                registered ORCID, ideally all authors have one
\author{Victor Pekkari}
\authornote{This work was conducted while the authors were at Neo4j.}
\email{victor.pekkari@gmail.com}

\author{Harry Wik}
\authornotemark[1]
\email{mail@harrywik.com}
\affiliation{%
  \institution{Lund University, Faculty of Engineering}
  \city{Lund}
  \country{Sweden}
}

\author{Alfred Clemedtson}
\affiliation{%
  \institution{Neo4j}
  \city{Malm{\"o}}
  \country{Sweden}
}
\email{alfred.clemedtson@neo4j.com}

\author{Borun Shi}
\affiliation{%
  \institution{Neo4j}
  \city{London}
  \country{England}
}
\email{brian.shi@neo4j.com}

\author{Xuan-Son Vu}
\affiliation{%
  \institution{Lund University, Faculty of Engineering}
  \city{Lund}
  \country{Sweden}
}
\email{xuan-son.vu@cs.lth.se}


\renewcommand{\shortauthors}{Pekkari et al.} %%MS: removed contents
%\renewcommand{\shorttitle}{} %%MS: added this one

\begin{abstract}
Graph Neural Networks on large graphs commonly rely on in-memory graph storage with ad hoc data processing pipelines for both training and inference. Graph databases provide scalable storage and efficient traversal for graphs that can evolve over time, but their applicability in GNN training and inference are rarely studied. We show that graph DBs can be used in GNN training and inference directly, executing sampling, feature retrieval, and even the full forward pass. We show different GNN training pipelines can perform on-the-fly sampling and retrieval in graph DBs using Cypher. We further provide optimisations that speed up the training process by \~2.5x while maintaining the same accuracy. We illustrate that fully in-DB inference is possible, independent of how the GNNs are trained. Our inference setup can be >10x faster than cold loading the graph into RAM with no extra memory overhead on real-world graphs. Our work lays the foundation for in-graph-DB GNN training and inference.
\end{abstract}


\maketitle

\section{Introduction}

Graphs are ubiquitous, with domains ranging from social networks to molecular structures. Predictive tasks on graphs include node classification and link prediction\cite{wu2020survey}. Graph Neural Networks (GNNs) have emerged as the dominant architecture for these tasks\cite{wu2020survey,velickovic2018gat}. Most GNNs produce node embeddings by aggregating information from local neighborhood for each node.

GNNs are particularly effective on data where relationships between entities carry important inductive bias. However, for a GNN of K layers, the computation graph of any node grows rapidly with K, a problem known as neighborhood explosion\cite{graphsage}. To address this issue, modern GNN training methods employ graph sampling that select subsets of full K-hop neighborhoods\cite{chen2018fastgcn, graphsaint, graphsage}. This introduces two additional stages into the training process, namely subgraph sampling on the topology and retrieval of feature vectors for every node in the sampled subgraph.

Graph database is an established technology for storing and retrieving graph data\cite{angles2008survey}. Many real-world industry-scale graphs are already stored in graph databases\cite{neo4j}. However, there is currently very little synergy between the field of GNNs and graph DB. Ad hoc storage and data processing pipelines are built for GNN training\cite{zhu2019aligraph, waleffe2022mariusgnnresourceefficientoutofcoretraining,Park_2022}, with few systems-level improvements for inference. Moreoever, even though many of these referenced work are inspired by techniques from databases, none of them directly utilise existing graph DBs in the training or inference pipeline.

We take an innovative, data-management perspective at GNN computation and formulate it as a database workload. Intuitively, most GNN forward equations are simply retrieving nodes according to some neighborhood (query) definition and perform aggregation. \cite{lopushanskyy2024gnns} touched upon this idea with a proof of concept. This inspires us to more tightly couple GNN training and inference process with graph DBs. Specifically, our contributions are:

\begin{itemize}
\item We show that graph DBs can be used for forward computation in GNN training, including sampling and feature retrieval. We note that the implementation can be done via either query (Cypher) or user-defined procedures (UDPs). We thoroughly benchmark different implementation options and provide optimisation (pre-aggregation) that greatly speed up training for certain classes of graphs and GNNs.
\item We show that fully in-DB inference is possible, independent of how the GNNs are trained. It is particular useful for real-world graphs already stored in graph DBs that may evolve over time. To the best of our knowledge, this is the first work that uses graph DBs as a full GNN inference system.
\end{itemize}

\section{Preliminary}
\newcommand{\red}[1]{\textcolor{red}{\textbf{#1}}}
\subsection{Graph Neural Networks}
A popular class of GNNs is message passing GNNs\cite{mpgnn-framework}.
At layer $k$, the representation $\boldsymbol{h}_v^{(k)}$ of node $v$ is computed as
\[
\mathbf{h}_v^{(k)} =
\phi^{(k)}\!\left(
\mathbf{h}_v^{(k-1)},
\bigoplus_{u \in \mathcal{N}(v)}
\psi^{(k)}\!\left(\mathbf{h}_v^{(k-1)}, \mathbf{h}_u^{(k-1)}, \mathbf{e}_{vu}\right)
\right)
\]\label{eq:mpnn}
where $\mathcal{N}(v)$ denotes some neighborhood of node $v$. 
The message function $\psi^{(k)}$ computes one message for each neighbor $u \in \mathcal{N}(v)$. $\mathbf{e}_{vu}$ is some edges feature on the edge $vu$.
The aggregation function $\bigoplus$ is usually some permutation-invariant multiset function (e.g mean, sum). 
Finally, the update function $\phi^{(k)}$ combines the previous representation $\mathbf{h}^{(k-1)}_v$ of node $v$ with the aggregated neighborhood representation and outputs the new embedding $\mathbf{h}^{(k)}_v$.

GNNs are typically trained and served from an in-memory representation of the graph\cite{lee2022smartsage, fey2019pyg, zheng2020distdgl, wang2023stag, wan2023g3}, with trade offs between the common formats. CSR and CSC support fast neighbor lookup during sampling\cite{mostafa2023fastsample}, but must be rebuilt whenever the graph changes \cite{islam2022vcsr}. COO avoids this cost, since new edges can simply be appended, but does not support fast sampling, as finding a node's neighbors requires scanning the entire edge list\cite{fey2019pyg}.
% \textbf{Training.} 
% Subgraph sampling requires many random look-ups in the adjacency matrix at the start of every batch, and keeping the graph on disk\textcolor{blue}{TODO: emphasize less about on-disk} would considerably slow this down compared to RAM.\textcolor{blue}{But if represented as CSR, it is no longer "random lookup on adjacency"?} 

\textbf{Training.} As graphs grow, distributed training is often used to remediate memory and throughput bottlenecks. The graph and features are partitioned across machines that synchronize weight updates. This improves throughput through parallelism, but at the cost of extra hardware, maintainence effort and graph partition time\cite{zheng2020distdgl, wang2019dgl}. Partitioned graphs additionally introduce extra memory (duplicate data) or slower querying\cite{Karypis1998AFA, wan2023g3}. Some performs sampling on the partitioned graph directly in GPUs, but optimising GPU utilization and overall throughput remains challenging\cite{yang2022wholegraph, liu2020G3}.

\textbf{Inference.} 
Inference does not require the same repeated random access that makes training RAM-bound, since predictions are typically needed on-demand rather than continuously. Keeping the full graph resident in RAM between requests is therefore unnecessarily expensive when low latency inference isn't crucial.
Keeping the graph on disk between inference requests could substantially lower the hardware cost, while providing acceptable inference latency.

There are several algorithmic improvements that make inference on large graphs more efficient, \cite{naman2025ripple} incrementally updates embeddings after graph changes rather than recomputing full aggregations, \cite{serving2023si} compresses graphs by reducing the number of nodes while minimising the approximation error.

\subsection{Graph Databases}

Graph databases provide native support for storing nodes and edges and their types and properties. Common graph databases include Neo4j\cite{neo4j}(disk-based) and K\`uzu\cite{chen2023kuzu}(in-memory).
On top of storage, there are query languages and query engines that efficiently traverse the graph. Cypher\cite{francis2018formal} is a common graph query language that follows the recent GQL standard\cite{gheerbrant2024gql}. The core concept of Cypher and GQL is graph pattern matching, where one can express patterns with filters on graphs in the DB, together with expressive transformations and aggregations. In addition to the declarative query languages, many graph DBs also support user-defined procedures (UDPs)\cite{neo4j-udp} that allow developers to explicitly control traversal and retrieval through the DB kernel.

\section{System Design}
This section outlines our approaches for shifting parts of the GNN forward equations into graph DBs for both training and inference.

\subsection{Training Pipelines}\label{sec:training_variants}
\begin{figure}[h]
    \centering
    \includegraphics[width=1\linewidth]{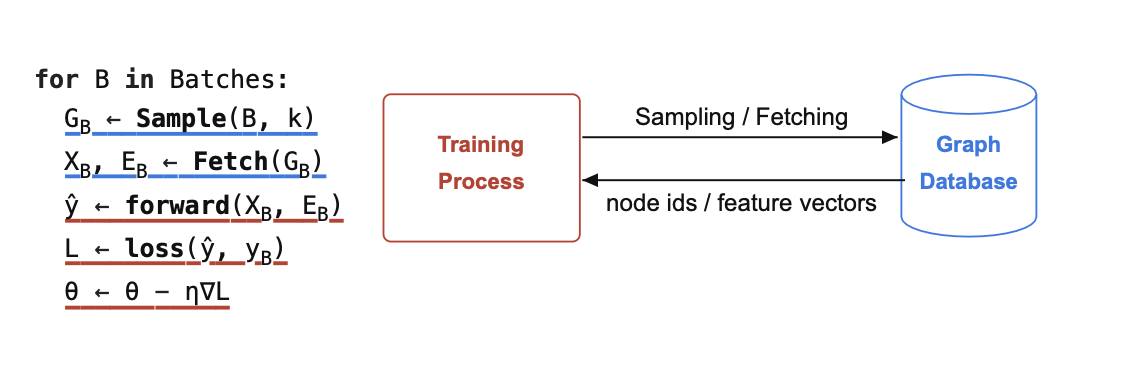}
    \caption{Illustration of computations that happen in the training process and the graph database when processing a batch. Note that the training process can live in the same or different machine to the graph DB.}
    \label{fig:gnn-batch-flow}
\end{figure}

Our graph database-backed training pipeline stores the graph in a graph
database while executing the learning algorithm in a separate training
process. For each mini-batch, the database performs subgraph sampling
directly on the graph on disk and returns the sampled subgraph and
associated node features. Since only
the sampled subgraph must reside in memory, this reduces the RAM
requirement of the training process. The returned features are then used
for the forward pass, loss calculation, and weight update, all executed in
the training process (Figure~\ref{fig:gnn-batch-flow}). 

We evaluate three increasingly optimized variants of the graph
database-backed training pipeline. We start from a baseline using declarative Cypher
queries for sampling and feature retrieval. Optimizations are then made to how computation is executed within the
DB and how much of the forward pass is moved into it. We compare training behaviour with standard in-memory PyG
implementation.

\textbf{In-memory reference.}
Reference implementation of GNN training in PyG with the graph stored in RAM. This implementation is used to verify that the graph database-backed pipelines have the same learning behavior.

\textbf{Cypher baseline.}
Our baseline implementation expresses both subgraph sampling and feature retrieval as declarative Cypher queries.

\textbf{UDP.}
Our second implementation replaces the Cypher queries with UDPs that are implemented in the DB.

% While the sampling algorithms considered in this work can be expressed in Cypher, doing so often requires complex declarative queries that introduce additional query planning, execution, and result construction overhead. In contrast, Java procedures provide fine-grained control over graph traversal and the construction of the returned data structures, allowing the sampling algorithms to be implemented more directly and efficiently while preserving the functionality of the training pipeline.

\textbf{Aggregation in the graph database (Pre-aggregation)}
Our final variant reduces data transferred between the graph database and the training process by performing the parameter-free neighborhood aggregation of the first GNN layer inside the database. Instead of returning feature vectors for every sampled node, the database returns the feature vectors required by the learnable parts of the model together with pre-aggregated neighborhood vectors for the outermost sampled nodes. This eliminates the need to transfer the feature vectors of nodes that only contribute to the first aggregation operation, reducing the amount of data exchanged between the database and the training process. The remaining learnable computations, including all weight multiplications and backpropagation, are still executed in the training process.

Formally, consider a message-passing layer
\[
\mathbf{h}_v^{(k)} = \phi_\theta^{(k)}\Big(\mathbf{h}_v^{(k-1)},
\bigoplus_{u \in \mathcal{N}(v)} \psi_\theta^{(k)}(\mathbf{z}_{vu})\Big).
\]
We call this layer \emph{exactly pre-aggregable} if the message function
$\psi_\theta^{(k)}$ can be split into a parameter-free part
$\Psi_{\mathrm{free}}^{(k)}$ followed by a learnable part $g_\theta^{(k)}$,
i.e.\ $\psi_\theta^{(k)} = g_\theta^{(k)} \circ \Psi_{\mathrm{free}}^{(k)}$,
such that $g_\theta^{(k)}$ commutes with the aggregation $\bigoplus$:
\[
\bigoplus_{u \in \mathcal{N}(v)}
g_\theta^{(k)}\big(\Psi_{\mathrm{free}}^{(k)}(\mathbf{z}_{vu})\big)
=
g_\theta^{(k)}\Big(\bigoplus_{u \in \mathcal{N}(v)}
\Psi_{\mathrm{free}}^{(k)}(\mathbf{z}_{vu})\Big).
\]
When this holds, the parameter-free
$\mathbf{a}_v^{(k)} = \bigoplus_{u \in \mathcal{N}(v)}
\Psi_{\mathrm{free}}^{(k)}(\mathbf{z}_{vu})$
can be computed entirely inside the database, and the learnable
transformation $g_\theta^{(k)}$ is applied afterward in the training
process, with no change to $\mathbf{h}_v^{(k)}$.

\subsection{Inference Pipelines} \label{sec:inference}
Inference does not require backpropagation. This implies the full forward pass can be executed either inside the graph database or in a separate inference process. As a result, it introduces several possible inference architectures that further moves the forward pass into the graph database\textit{}. We evaluate four such architectures to understand the trade-offs between latency and infrastructure cost.

% We evaluate four inference architectures that differ in where the graph is stored, where the forward pass is executed, and how the graph database execution is defined (imperatively or declaratively). The variants are described below.

\textbf{In-memory reference.}
This baseline has two versions, one (Cold) that loads the entire graph from disk to RAM, and one (Hot) that already has the graph in RAM upon the inference request. The full forward pass is executed in a inference process running on a CPU without the use of any graph database.

\textbf{Graph database with external inference process (Neighborhood sampling).}
This variant stores the graph in a graph database while executing the forward pass in a separate inference process. For each inference request, the inference process first executes sampling and feature retrieval followed by the forward pass executed in the inference process.

\textbf{In-database inference with UDP.}
In this variant, the trained GNN is deployed inside the graph database as a UDP. The database performs both neighborhood retrieval and the complete forward pass, returning only the final predictions. This minimizes data transfer between the database and an external process.

\textbf{In-database inference with Cypher.}
The full forward pass and sampling are executed entirely within the graph database as a single declarative query. We provide two implementations with slightly different Cypher queries. In the first, matrix multiplicatoins are expressed with the general-purpose \texttt{reduce()} function, available in all Cypher versions. In the second, which we refer to as \texttt{in\_db\_cypher\_opt}, we use Cypher 25's native vector operations, which allow the aggregation to be expressed more directly.

It is important to reiterate that to the best of our knowledge, there is no prior work that studies fully in-DB inference of GNN models, despite the simplicity of the idea and a basic implementation (just some Cypher query). All prior work treat graph database simply as storage from which data is exported.

% The pipelines were monitored during the experiments, collecting the following metrics:

% \begin{itemize}
%     \item \textbf{Validation accuracy} per epoch during training,
%     \item \textbf{Latency of query subphases}: sampling latency, feature retrieval latency, serialization / deserialization and data assembly in Python,
%     \item \textbf{Python profiling}: time spent in forward passes, backward passes, taking steps with the optimizer as well as other Python functions,
%     \item \textbf{Sampling distributions} over edges and nodes included,
%     \item \textbf{System utilization}: RAM, CPU.
% \end{itemize}

% Additionally, for the inference experiments, we also collected the latency per batch of seed nodes processed as well as the performance on unseen test data.

\section{Experiments}
\subsection{Setup}

All experiments were run on a single machine with 32 vCPUs and 256~GB of RAM, running Ubuntu 22.04 LTS on an Intel x86-64 platform. The Neo4j databases were hosted on four NVMe drives unified into a single RAID~0 partition, providing high sequential and random I/O throughput for disk-resident graph access. Python~3.12 with PyTorch~2.7.1 and PyTorch Geometric (PyG)~2.7 is used. 
UDPs are written in Java for Neo4j.
For the experiments the following datasets were used: Cora~\cite{yang2016revisiting}, CoAuthor Physics~\cite{shchur2018pitfalls}, arXiv, ogbn-products and ogbn-papers100M~\cite{hu2020ogb}. More dataset statistics and model configuration are specified in Table \ref{tab:stats}. The hyperparameters were kept fixed throughout the experiments. For all training and inference pipelines that use Neo4j, we store the properties as byte arrays, which are more efficient during data transfer and decoding (Appendix ~\ref{app:byte}).

\begin{table}[H]
\centering
\scriptsize
\setlength{\tabcolsep}{3pt}
\begin{tabular}{@{}lrrrrr@{}}
\hline
 & \textbf{Cora} & \textbf{CoAuthor}& \textbf{arXiv}& \textbf{products} & \textbf{papers100M} \\
\hline
\# nodes & 2,708 & 34,493 & 169,343 & 2,449,029 & 111,059,956 \\
\# edges & 10,556 & 495,924 & 1,166,243 & 61,859,140  &  1,615,685,872 \\
Avg degree & 3.90 & 14.38 & 6.89 & 25.26 &  14.55 \\
Training nodes & 140 & 20,696 & 90,941 & 196,615 & 1,207,179 \\
Validation nodes & 500 & 3,449 & 29,799 & 39,323 &  125,268\\
Test nodes & 1,000 & 10,348 & 48,603 & 2,213,091 &  214,338\\
Embedding dimension & 1433 & 8,415 & 128 & 100 & 128 \\
1st hidden dimension & 12 & 32 & 64 & 64 & 192 \\
2nd hidden dimension & 12 & 32 & 48 & 64 & 192 \\
\shortstack[l]{Categories /\\Output dimension} & 7 & 5 & 40 & 47 & 172\\
\hline
\end{tabular}
\caption{Dataset statistics and training configuration.}
\label{tab:stats}
\end{table}

\begin{figure}
    \centering
    \includegraphics[width=1.0\linewidth]{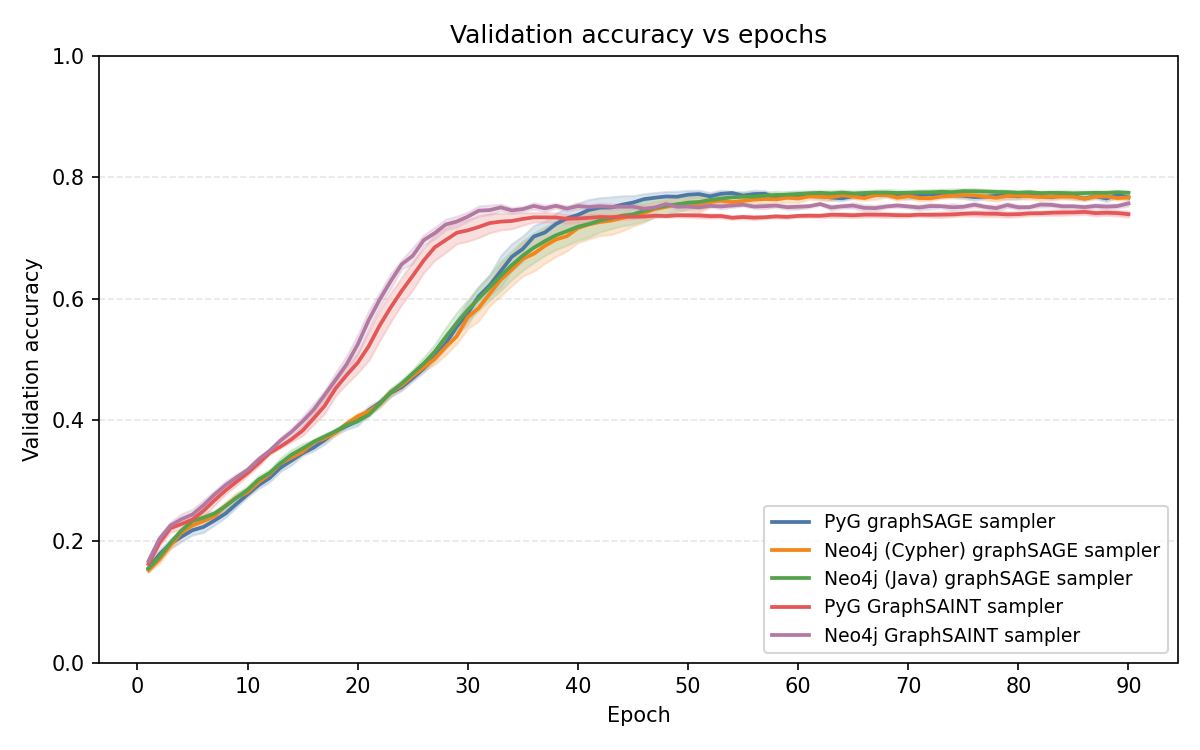}
\caption{Validation accuracy on Cora for GraphSAGE and GraphSAINT sampling implemented in PyG, Cypher and UDP. Each line shows the mean validation accuracy over 10 runs, and the shaded region shows the 95\% interval.}
\label{fig:sampler_acc_epoch}
\end{figure}

We validate our database-backed implementations against the corresponding pure PyG implementations using KL divergence, learning curves, and final predictive metric. 
KL divergence between nodes distributions and edge distributions are below 0.01 for GraphSAGE and GraphSAINT samplers in both Cypher and UDPs.
The validation accuracy converges similarly for all implementations as shown in Figure~\ref{fig:sampler_acc_epoch}. They prove various implementations are correct and subsequent time/memory systems-level comparisons are made for comparable training performance.

\subsection{Training}
% To identify where optimization efforts should be focused, we profile the graph database-backed training pipeline by measuring the latency of its individual stages. In particular, we separate subgraph sampling, feature retrieval, and GNN compute. We further break down the communication between the graph database and training process into graph database execution, data transfer, and ETL. We do this by profiling the Cypher query, and timing the ETL and total query time.

\begin{figure}
    \centering
    \includegraphics[width=\linewidth]{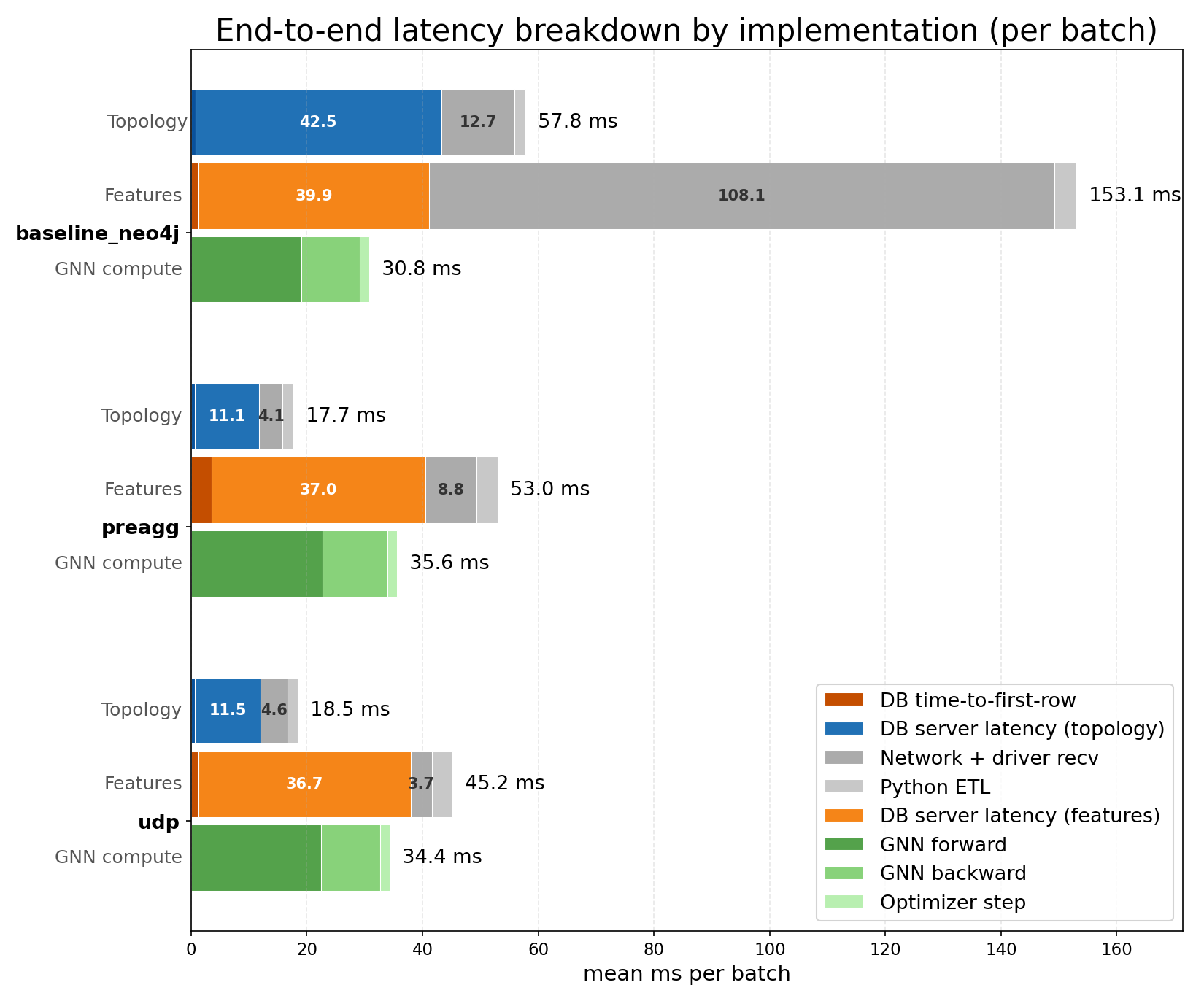}
    \caption{\texttt{ogbn-papers100M}, GraphSAGE-sampling. Breakdown of time spent in training. \texttt{baseline\_neo4j} represents Cypher implementation. Using UDP drastically reduces both data transfer and sampling execution time. Using UDF speeds up \texttt{baseline\_neo4j} by \~2.5 times. Using pre-aggregation does not provide improvement in latency for the when feature dimension is low in this case.}
    \label{fig:papers100M_training_optimizations}
\end{figure}

Figure \ref{fig:papers100M_training_optimizations} compares the execution time of each pipeline stage across all training pipeline variants. The results show that the baseline pipeline supported by the graph database is dominated by the sample-fetch phase, motivating optimizations that reduce query execution time, or data transfer time.

\subsubsection{Sampling: Cypher vs UDP}
\begin{figure}
    \centering
    \includegraphics[width=\linewidth]{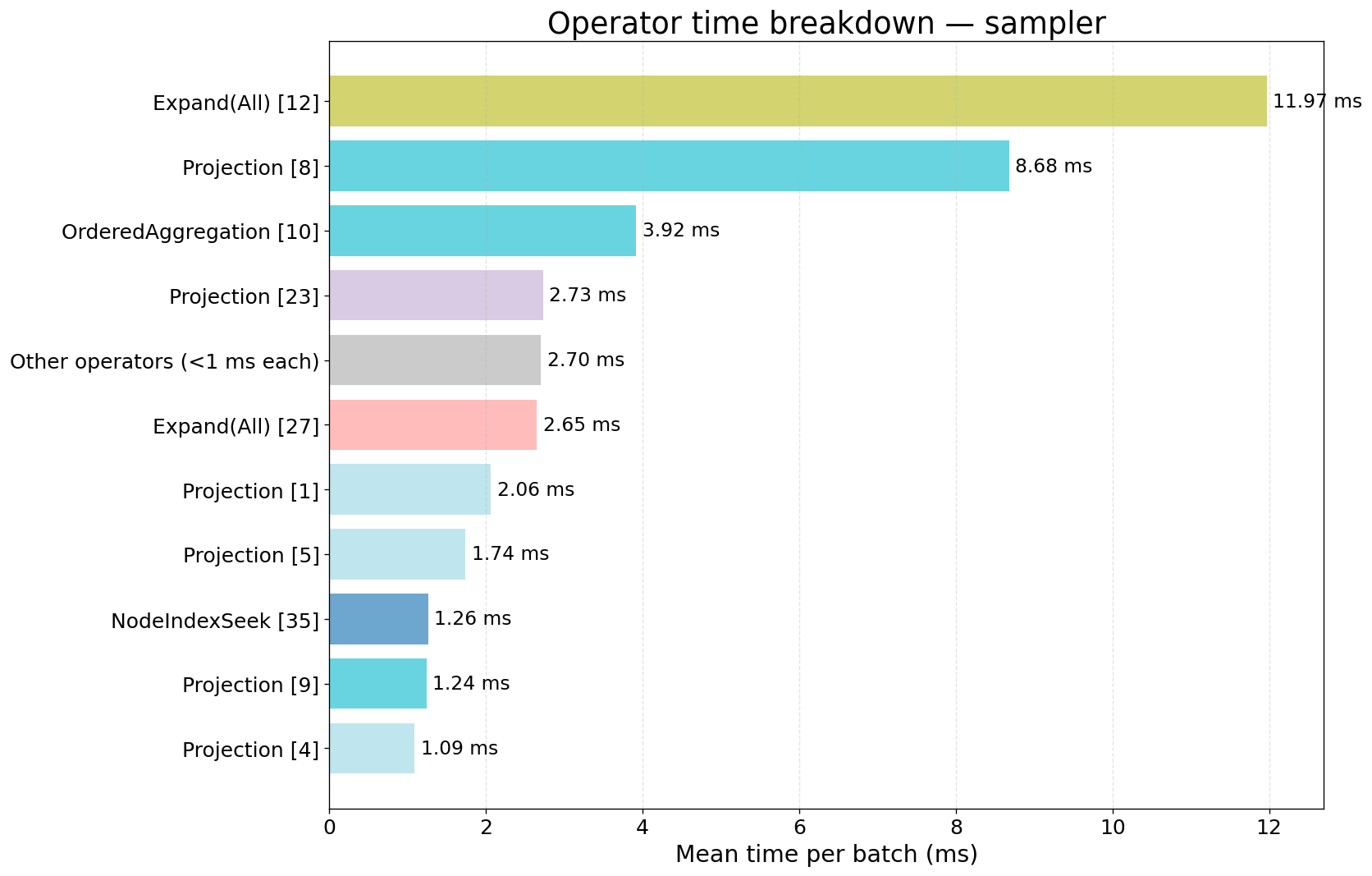}
    \caption{Per-operator execution-time breakdown of the Cypher GraphSAGE sampler on \texttt{ogbn-papers100M}. Graph traversal (\texttt{Expand(All)}) accounts for roughly $14.6$~ms, less than the combined cost of \texttt{Projection} and \texttt{OrderedAggregation},
which only build and sort intermediate lists rather than touch the graph.
}
    \label{fig:baseline_cypher_db_operators_latency}
\end{figure}

\begin{figure}
    \centering
    \includegraphics[width=\linewidth]{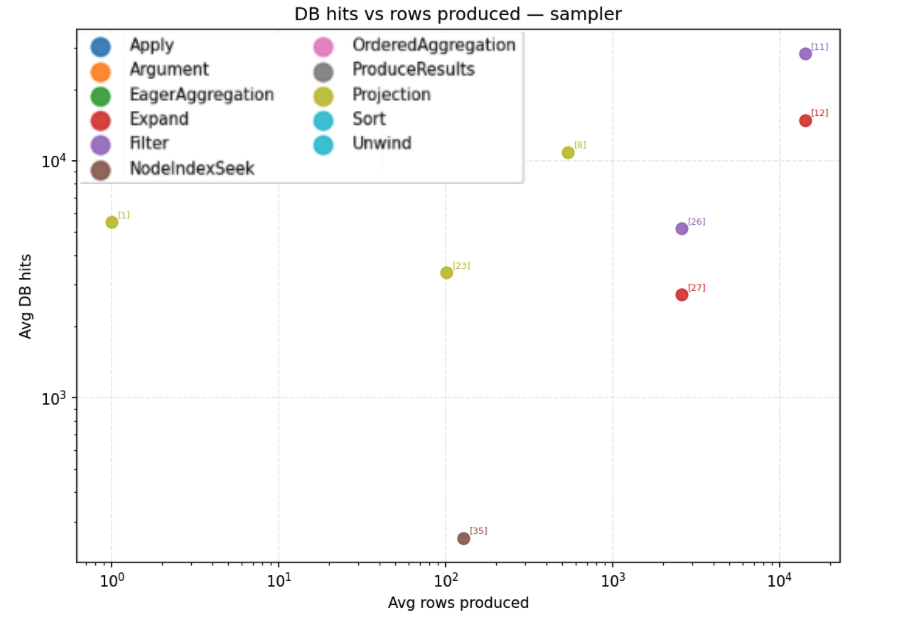}
\caption{Database hits vs. rows produced per operator for the GraphSAGE sampling-query on \texttt{ogbn-papers100M} (log-log, y-axis
starts at 100). The \texttt{Expand} operator produces over 10,000 rows,
well above the necessary $128 \cdot (1 + 10 + 10\cdot5) =
7808$ nodes, for fanout \texttt{[10, 5]} and \texttt{batch\_size=128}, showing that
the database fetches raw relationships later discarded during sampling.}
    \label{fig:DB-rows}
\end{figure}

It can be seem from Figure \ref{fig:papers100M_training_optimizations} that UDP is faster than the Cypher counterpart. The query profile for the Cypher sampling in Figure~\ref{fig:baseline_cypher_db_operators_latency} indicates that graph traversal is the primary cost of the sampling query, with \texttt{Expand(All)} accounting for the largest share of execution time as it retrieves neighboring vertices for each sampled node. Although this graph traversal operator is essential for sampling, Figure \ref{fig:DB-rows} shows that the number of expanded nodes exceeds the number of sampled nodes, indicating that the query expands more nodes than necessary. However, a significant proportion of the runtime is also spent in \texttt{Projection} and \texttt{OrderedAggregation}, which construct intermediate results and group neighbors into per-node collections. These operators do not contribute directly to the sampling logic but instead represent the overhead of the declarative query execution model, where intermediate tuples are materialized, transformed, and aggregated between stages. In contrast, a UDP implementation using reservoir sampling\cite{vitter1985reservoir} performs neighbor selection incrementally during traversal, requiring only $O(k)$ additional memory for a sample of size $k$, rather than materializing all $d$ neighbors. Although both approaches have $O(d)$ time complexity with respect to node degree $d$, the Cypher implementation incurs additional processing overhead due to intermediate data management. This trend is reflected in our experimental results, where on the \textit{ogbn-papers100M} dataset (average degree 15), the UDP sampler achieved a 3.8x speedup over the Cypher implementation, while on the denser \textit{ogbn-products} dataset (average degree 25), the speedup increased to 5.2x. The performance of the Cypher-based sampler could be improved by reducing the number of candidate neighbours processed during the \texttt{Expand(All)} stage. If the query could restrict expansion to a smaller subset of candidate neighbours before materialisation, fewer intermediate tuples would be created, reducing the overhead of the subsequent \texttt{Projection} and \texttt{OrderedAggregation} operators. In Appendix ~\ref{app:lantency-ogbn-products}, we show that for dense graphs, the topology traversal time are significantly reduced in preaggregation and UDP.

\subsubsection{Fetching: Cypher vs. UDP}

Feature fetching latency is also reduced when switching from Cypher to UDP. However, as shown in Figure~\ref{fig:papers100M_training_optimizations}, the largest improvement is not in graph database execution time, but in data transfer time. This is likely because the Cypher query used to retrieve feature vectors is relatively simple and therefore already executes efficiently. Instead, the primary opportunity for optimization lies in how the fetched data is represented and transferred. The UDP implementation allows all feature vectors to be serialized into a single byte array before transmission, significantly reducing the per-feature transfer overhead compared to transferring each feature vector as an individual record. As a result, the reduction in communication overhead contributes more to the overall performance improvement than the reduction in query execution time.

\subsubsection{Aggregation in the graph database (Pre-aggregation)}
% Moving parameter-free neighborhood aggregation into the graph database changes where part of the forward pass is executed. We therefore compare the training convergence of the pre-aggregation implementation (Section \ref{sec:preaggregation}) with its corresponding in-memory implementation. When running training, we found no statistically significant difference between the pre-aggregation implementation and the in-memory system at any epoch.

% % Figure \ref{fig:preagg_training_convergence} shows nearly identical learning curves. 

% Combined with the previously verified sampling implementation (Section \ref{sec:learning_eq}), this indicates that pre-aggregation preserves the intended learning behavior.

\begin{figure}
    \centering
    \includegraphics[width=0.8\linewidth]{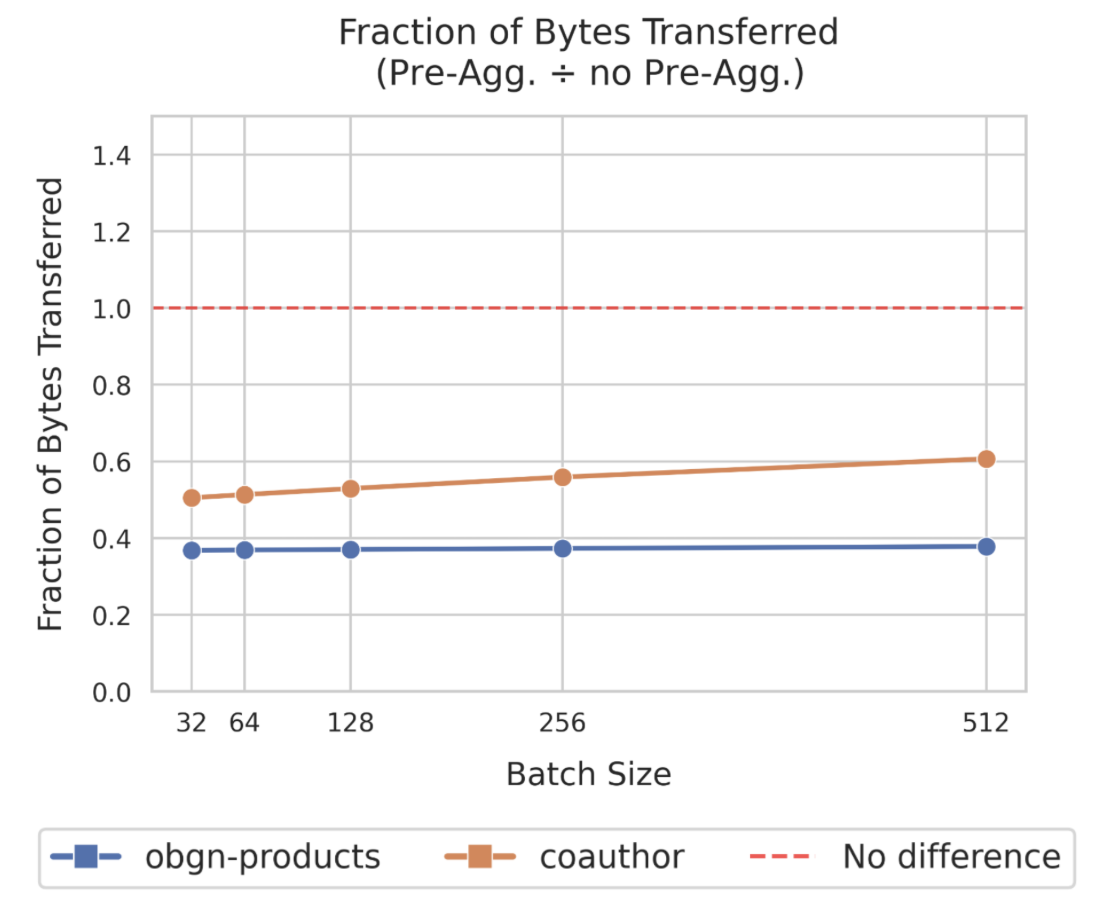}
\caption{GraphSAGE-sampling. Pre-aggregation reduces data sent from the DB to training process between ca. 40\% - 60\% for Fanout \texttt{[10, 5]}.}
    \label{fig:preagg_10_5}
\end{figure}

\begin{figure}
    \centering
    \includegraphics[width=\linewidth]{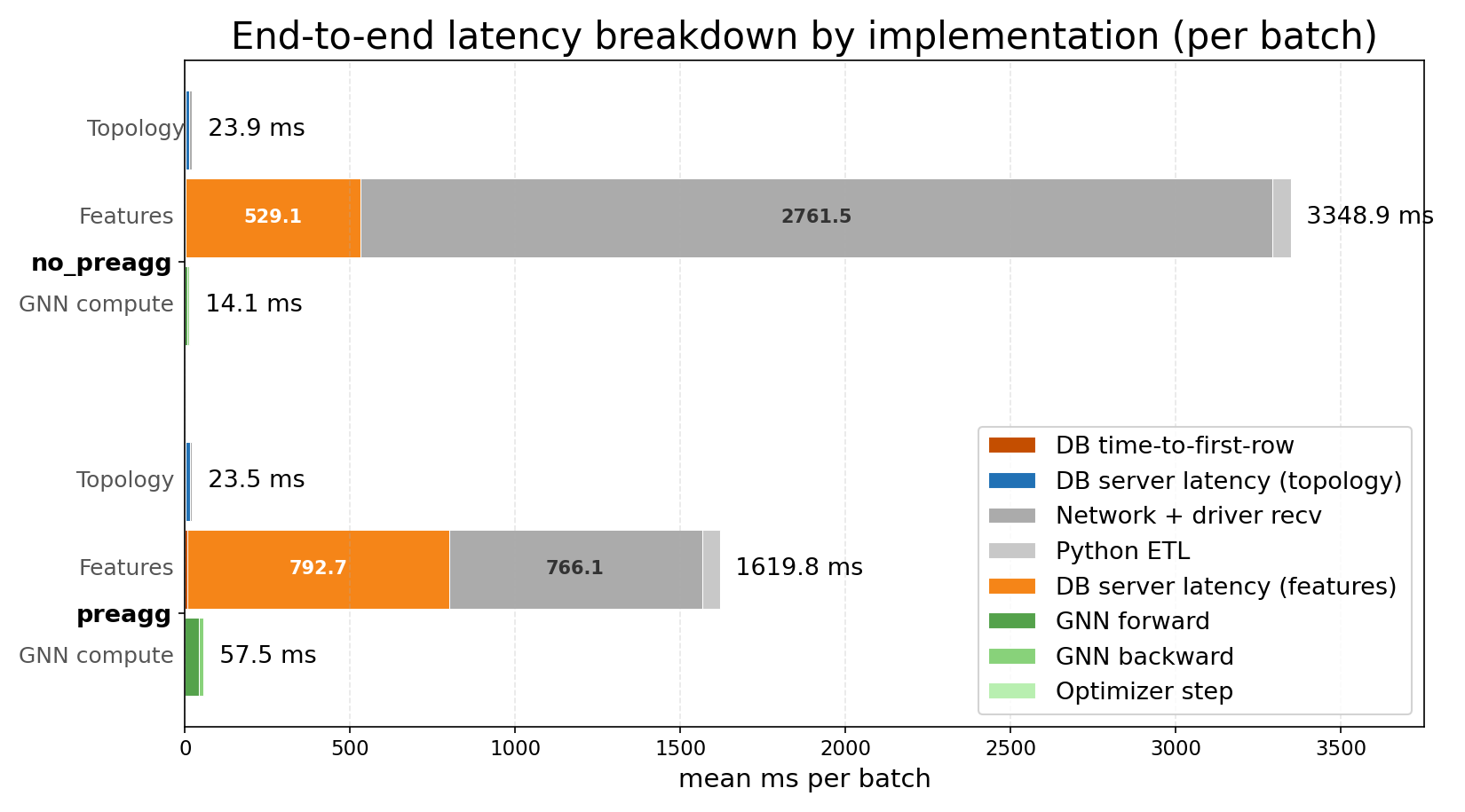}
    \caption{\texttt{CoAuthor Physics}. GraphSAGE-sampling. Having the highest feature dimensionality, this dataset benefits most from pre-aggregation, as the added DB-side compute and querying cost is outweighed by lower \texttt{Network + driver recv} time from sending less data.}
    \label{fig:coauthor-preagg-latency}
\end{figure}

Moving the aggregation operation of the first message passing layer to the graph database reduced the amount of data sent for all 5 datasets. Figure \ref{fig:preagg_10_5} shows the reduction for two of them. It did not lead to a feature-fetching speed-up for all datasets though. Figure \ref{fig:papers100M_training_optimizations} shows that the UDP implementation is fetching the features faster the pre-aggregation implementation for the \textit{ogbn-papers100M} with feature vectors of 100 dimensions. Figure \ref{fig:coauthor-preagg-latency} shows for the coauthor dataset with feature dimension of 8415, feature-fetching latency was more than halved when using pre-aggregation. 
We conclude that pre-aggregation always reduces the amount of data sent, which further reducing the memory requirements of the training process.
However, it only leads to a speedup for datasets with high dimensional vectors, due to the additional computation needed in the DB.

% \begin{figure}
%     \centering
%     \includegraphics[width=\linewidth]{figures/preagg_equivalence.png}
%     \caption{Training convergence on Cora for the same GCN architecture, comparing standard graph stored in-memory PyG execution with a pre-aggregation pipeline where first-layer neighborhood summaries are computed in Neo4j and consumed by a modified PyG forward pass.}
%     \label{fig:preagg_training_convergence}
% \end{figure}

\subsection{Inference}
\begin{figure}
    \centering
    \begin{subfigure}{\linewidth}
    \centering
    \includegraphics[width=\linewidth]{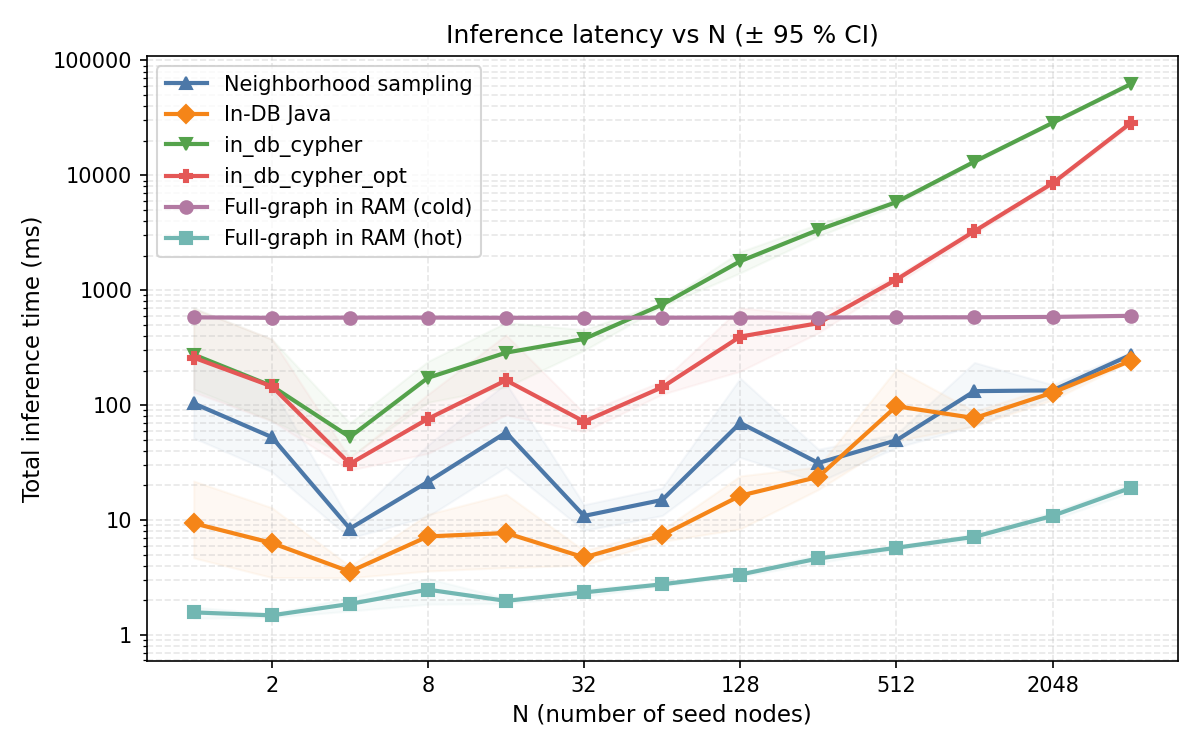}
    \caption{\texttt{ogbn-arxiv}. GraphSAGE-sampling four different implemented inference variants using the graph DB compared in mean batch processing latency against in-memory inference with the graph `hot' in RAM and `cold' -- firstly loading it to RAM before inference. N on x-axis represents the number of nodes we classify. See Section~\ref{sec:inference} for descriptions of each inference variant.}
    \label{fig:arxiv-inferenc-latencies}
    \end{subfigure}
    \begin{subfigure}{\linewidth}
    \centering
    \includegraphics[width=\linewidth]{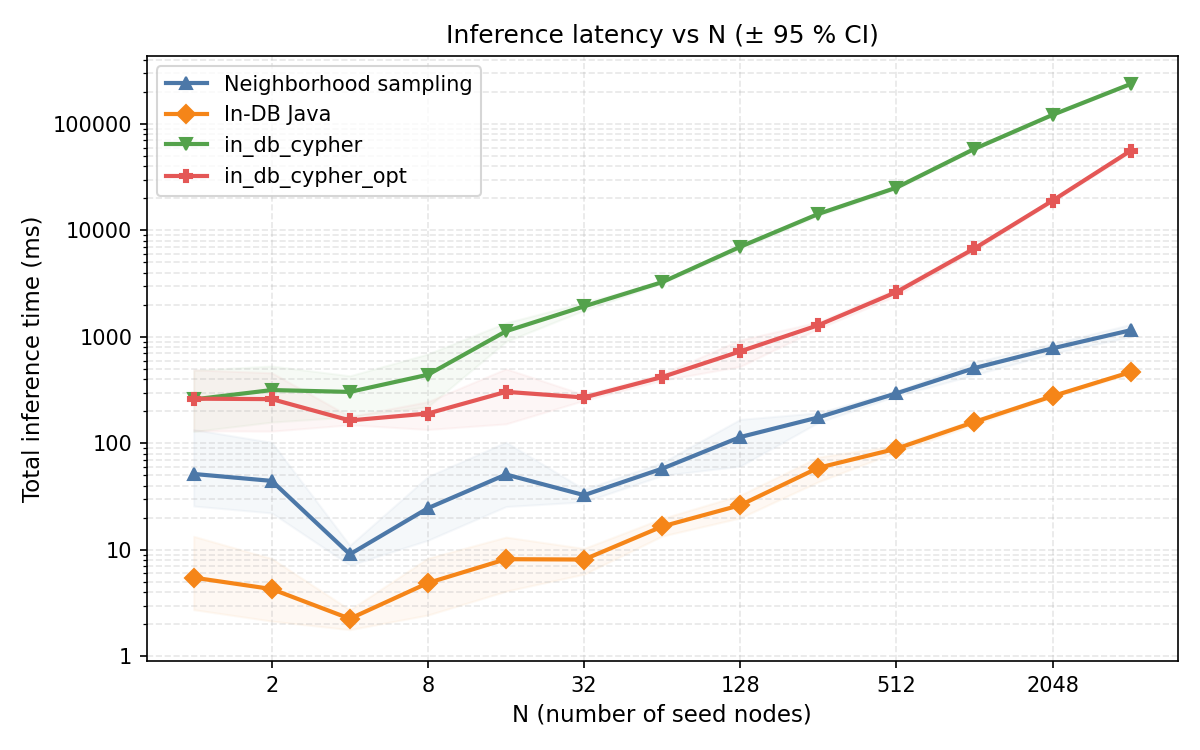}
    \caption{\texttt{ogbn-papers100M}. Benchmarks of four different inference variants behaves similarly on the different graph.}
    \label{fig:ogbn-inferenc-latencies}
    \end{subfigure}
    \caption{Benchmark of inference in databases.}
\end{figure}

We implement and benchmark the 5 inference strategies outlined in Section~\ref{sec:inference}. For any GNN model, we can export a definition of it (for example in torch state\_dict) and translate that to Cypher or UDP, with the model weights as parameters. This means the models can be trained anywhere with any library.

We also include the time for each implementation.
As shown in Figure~\ref{fig:arxiv-inferenc-latencies}, having the graph already in RAM and operating on it in PyG is the fastest across batch sizes. This target baseline assumes the graph remains static.
If an initial load of the graph into RAM from disk has to be performed (purple), it is slower then the UDP implementation of in-DB inference (orange), up to at least for 4096 nodes for this graph. 
Similar to what we observe during training, in-DB Cypher implementations are again inferior than in-DB UDP implementations. The intuitive interpretation is that UDPs (an imperative program written in Java against the kernal API in Neo4j) offers fine control of data access and computation, which can be seem as a specialised pre-compiled GNN query, compared to Cypher, whose query planner is designed for amortised optimal plan for arbitrary queries and graphs.

in-DB inference also uses minimal memory as it only needs to materialise the inference nodes with one layer of frontier nodes, leaving the remainder of the graph on disk. Our in-DB UDP implementation for 2048 nodes finish in <1s with 32GB RAM, and \textasciitilde 0.1s with 128GB RAM, when full graph PyG loading OOM.

Importantly, the advantageous in-DB UDP (orange) implementation implies that extremely efficient inference for graphs stored in DBs in real-world industry settings is now possible. A user can perform inference on any set of nodes in the DB with our in-DB UDP to obtain new emebddings or classifications.

\section{Conclusion}
\textbf{Limitations and Future Work.} 
Exact time and memory resource differences of various in-DB training and inference implementations will vary for other non-Neo4j disk-base graph DBs. However, the general trend and observations will be valid.
We only included the well-known GraphSAGE (node-wise) and GraphSAINT (subgraph-based) sampling techniques. Other approaches such as layewise samplings are not studied. Other distributed training systems or GPUs are not included. They are orthogonal improvements that can be incorporated.

\textbf{Summary.} In this work, we focused on the current lack of overlap between the field of GNNs and graph DB. We take an innovative data-management perspective to GNN training and inference, proposing that the majority of the computation (forward equations) can be shifted into graph DBs. We ask the question of whether an operational graph DB can be used as the execution engine of a GNN and confirm positively. We also noted that computations inside graph DBs can be achieved with queries (Cypher) or UDPs. We provided several implementations of training and inferences and benchmarked the runtime behaviour thoroughly. We find that training with on-the-fly retrieval in graph DBs can be achieved with Cypher, and that it can be significantly optimised with UDPs and pre-aggregation (under high-dimension embeddings). We also find that the most efficient inference can be achieved with in-DB UDP implementation of GNNs, even on graphs that can constantly change. This tight coupling of GNNs and graph DB at inference time is previously unstudied. Our work provides new training and inference systems for GNNs and use case scenarios for graph DBs.

\section*{Artifacts}
All our code, including the Cypher and UDP implementations of different GNN layers, are available on \url{https://github.com/harrywik/neo4j-mpgnn-thesis}. Detailed setup and reproducibility scripts, and on how to extend our work, are described in the README.md.

\appendix
\section{Neo4j embedding storage format}
\label{app:byte}

\begin{figure}
    \centering
    \includegraphics[width=\linewidth]{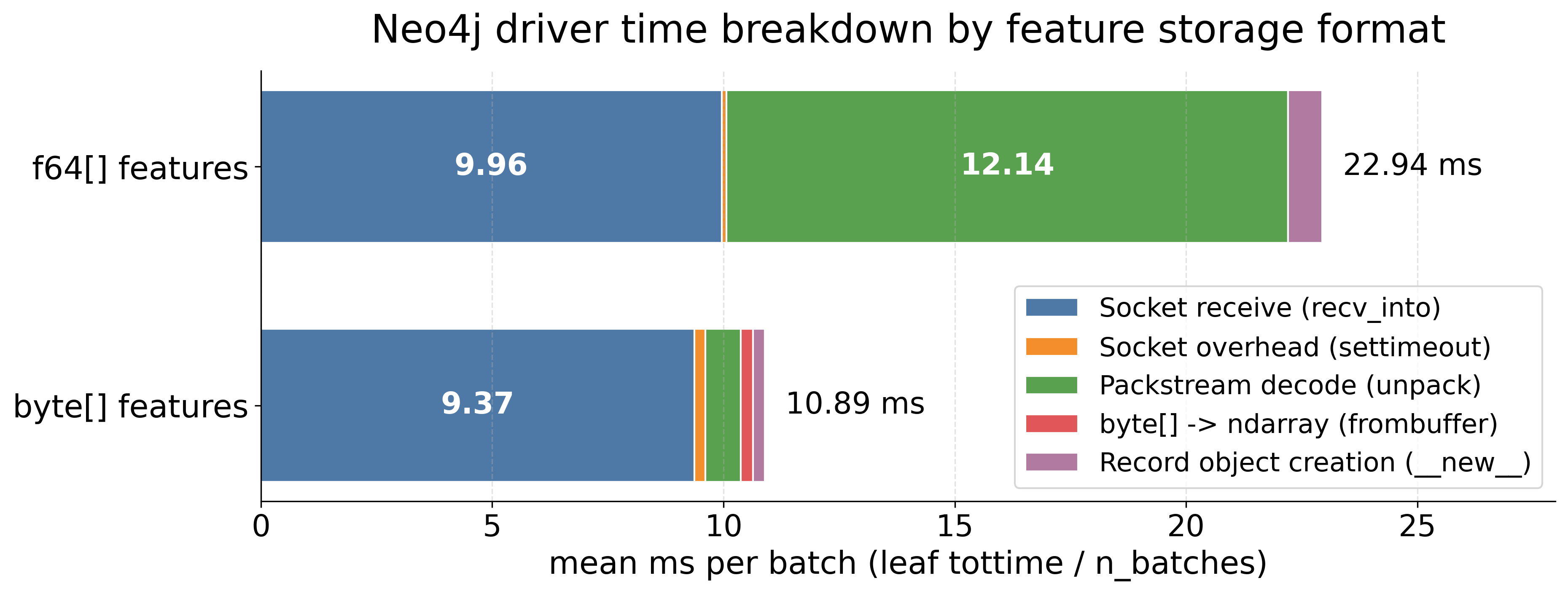}
    \caption{Representing embedding properties as \texttt{byte[]} reduces the time taken by driver by more than half, compared to \texttt{float64[]} format, due to the large reduction in the packstream decode operation in the Bolt client driver.}
    \label{fig:bolt_time_decomposition}
\end{figure}

Casting embeddings into compute-ready tensors does not incur a large overhead when storing and transmitting them as \texttt{byte[]} properties, as shown in Figure \ref{fig:bolt_time_decomposition}. Conversely, if embeddings are stored in Neo4j as \texttt{f64[]}, significant overhead is introduced. This is due to the type conversion required because PyG expects \texttt{f32} elements, whereas Neo4j stores \texttt{f64}, resulting in a doubling of the size of the TCP-payload. Even more critical is the behavior of the Bolt client driver, which casts every incoming \texttt{f64} into a Python \texttt{float}, creating a substantial overhead. The Python Global Interpreter Lock (GIL) and object instantiation during a linear scan of a larger array incurs high cost compared to the bulk \texttt{memcpy} of a byte buffer that can be performed when using \texttt{byte[]} in transit. This optimization is used for all setups including the baseline and the effects on the Cora dataset is shown in Figure \ref{fig:bolt_time_decomposition}. 

\section{Breakdown of training time on ogbn-product}
\label{app:lantency-ogbn-products}
\begin{figure}
    \centering
    \includegraphics[width=\linewidth]{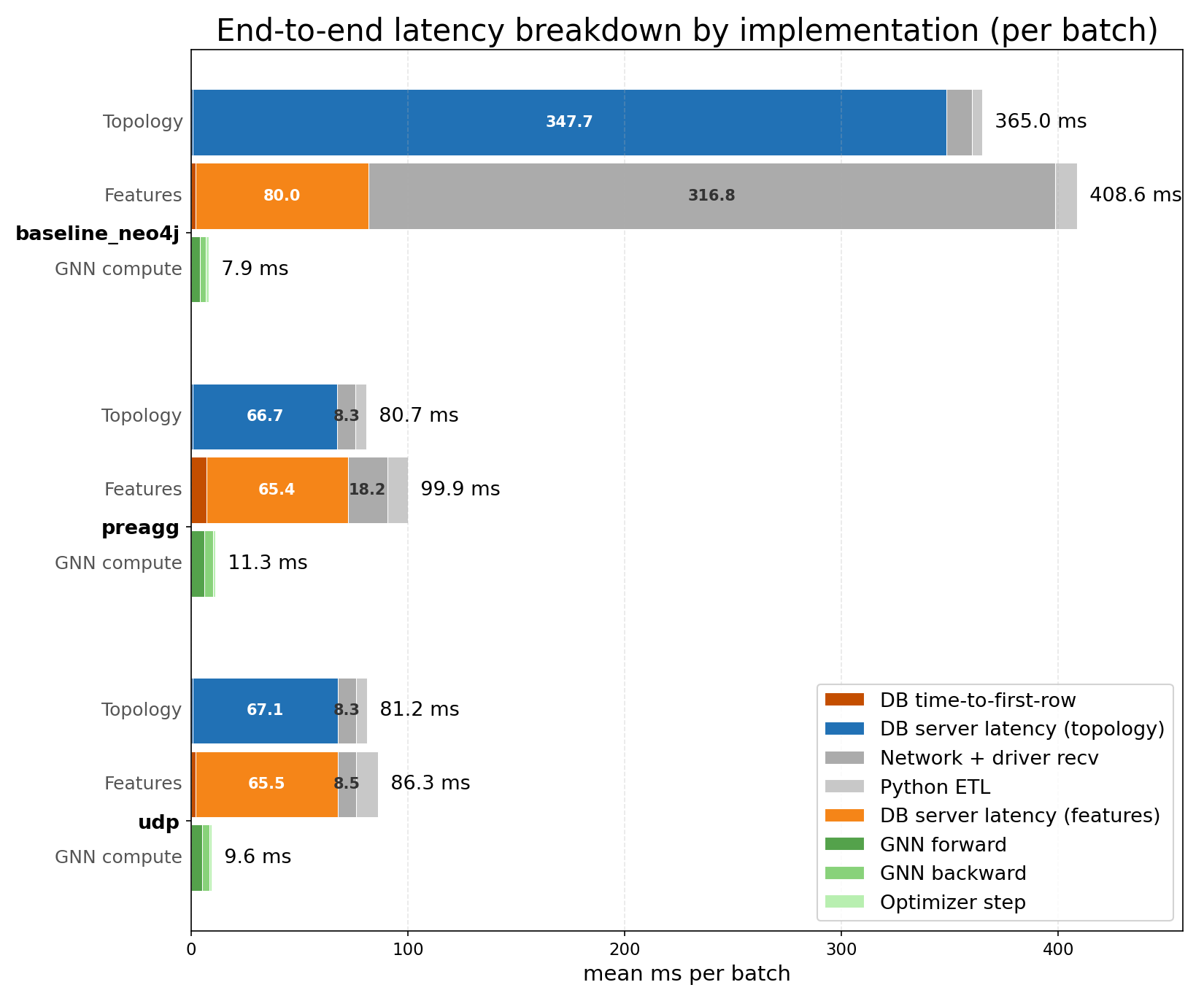}
    \caption{\texttt{ogbn-products}. Breakdown of time spent in training.}
    \label{fig:products_training_optimizations}
\end{figure}

In addition to Figure~\ref{fig:products_training_optimizations}, we provide the breakdown of training time of a large and dense graph. The time taken by graph traversal for \texttt{baseline\_neo4j} is significantly higher.

% \begin{acks}
% \end{acks}

\balance
\bibliographystyle{ACM-Reference-Format}
\bibliography{sample-base}


\end{document}